\documentclass{article}
\usepackage[utf8]{inputenc}

\usepackage{booktabs}
\usepackage{xcolor}
\usepackage{bbm}
\usepackage{listings}
\lstdefinestyle{Python}{
    language        = Python,
    basicstyle      = \ttfamily,
    keywordstyle    = \color{blue},
    keywordstyle    = [2] \color{teal}, % just to check that it works
    stringstyle     = \color{green},
    commentstyle    = \color{red}\ttfamily
}

\title{Captum: A unified and generic model interpretability library for PyTorch}
\author{Narine Kokhlikyan\thanks{Equal Contribution}, Vivek Miglani\footnotemark[1], Miguel Martin, \\
Edward Wang, Bilal Alsallakh, Jonathan Reynolds,\\
Alexander Melnikov, Natalia Kliushkina,  \\
Carlos Araya, Siqi Yan, Orion Reblitz-Richardson \\
Facebook AI \\
\{narine, vivekm, miguelmartin, hack, bilalsal, \\
jonreynolds, sanekmelnikov, natalial, caraya, \\
siqi, orionr\}@fb.com}
\date{}

\usepackage{graphicx}

\begin{document}

\maketitle

\begin{abstract}
In this paper we introduce a novel, unified, open-source model interpretability library for PyTorch~\cite{Paszke2019PyTorchAI}.
The library contains generic implementations of a number of gradient and perturbation-based attribution algorithms, also known as feature, neuron and layer importance algorithms, as well as a set of evaluation metrics for these algorithms. It can be used for both classification and non-classification models including graph-structured models built on Neural Networks (NN).
In this paper we give a high-level overview of supported attribution algorithms and show how to perform memory-efficient and scalable computations. 
We emphasize that the three main characteristics of the library are multimodality, extensibility and ease of use. Multimodality supports different modality of inputs such as image, text, audio or video. Extensibility allows adding new algorithms and features. The library is also designed for easy understanding and use. Besides, we also introduce an interactive visualization tool called Captum Insights that is built on top of Captum library and allows sample-based model debugging and visualization using feature importance metrics.


\end{abstract}

\section{Introduction}
\label{introduction}
Given the complexity and black-box nature of NN models, there is a strong demand for clear understanding of how those models reason. Model interpretability aims to describe model internals in human understandable terms and is an important field of Explainable AI \cite{8631448}. While building interpretable models is encouraged, many existing state-of-the-art NNs are not designed to be interpretable, thus the development of algorithms that explain black-box models becomes highly desirable. This is particularly important when AI is used in domains such as healthcare, finance or self-driving vehicles where establishing trust in AI-driven systems is critical. 

Some of the fundamental approaches that interpret black-box models are feature, neuron and layer importance algorithms, also known as attribution algorithms. Existing frameworks such as DeepExplain \cite{Ancona2018TowardsBU}, Alibi \cite{alibi} and InterpretML \cite{nori2019interpretml} have been developed to unify those algorithms in one framework and make them accessible to all machine learning model developers and practitioners. These frameworks, however, have insufficient support for PyTorch models. In Captum, we provide generic implementations of a number of gradient and perturbation-based attribution algorithms that can be applied to any PyTorch model of any modality. The library is easily extensible and lets users scale computations across multiple GPUs and handles large-sized input by dividing them into smaller pieces, thereby preventing out of memory situations.

Another important aspect is that both qualitative and quantitative evaluation of attributions are difficult. Visual explanations can be misleading \cite{sanity_checks} and evaluation metrics are subjective or domain specific \cite{yang2019benchmarking}. To address these issues, we provide generic implementations of two evaluation metrics called infidelity and max-sensitivity proposed in \cite{Yeh2019OnT}. These metrics can be used in combination with any PyTorch model and most attribution algorithms.

Lastly, model understanding research largely focuses on the Computer Vision (CV) domain whereas there are many unexplored NN applications that desperately need model understanding tools. Adapting CV-specific implementations for those applications is not always straightforward, thus the need for a well-tested and generic library that can be easily applied to multiple domains across research and production.

\section{An Overview of the Algorithms}
\label{supported_algorithms}
\begin{figure}[ht]
\vskip 0.2in
\begin{center}
\centerline{\includegraphics[width=1.0\columnwidth]{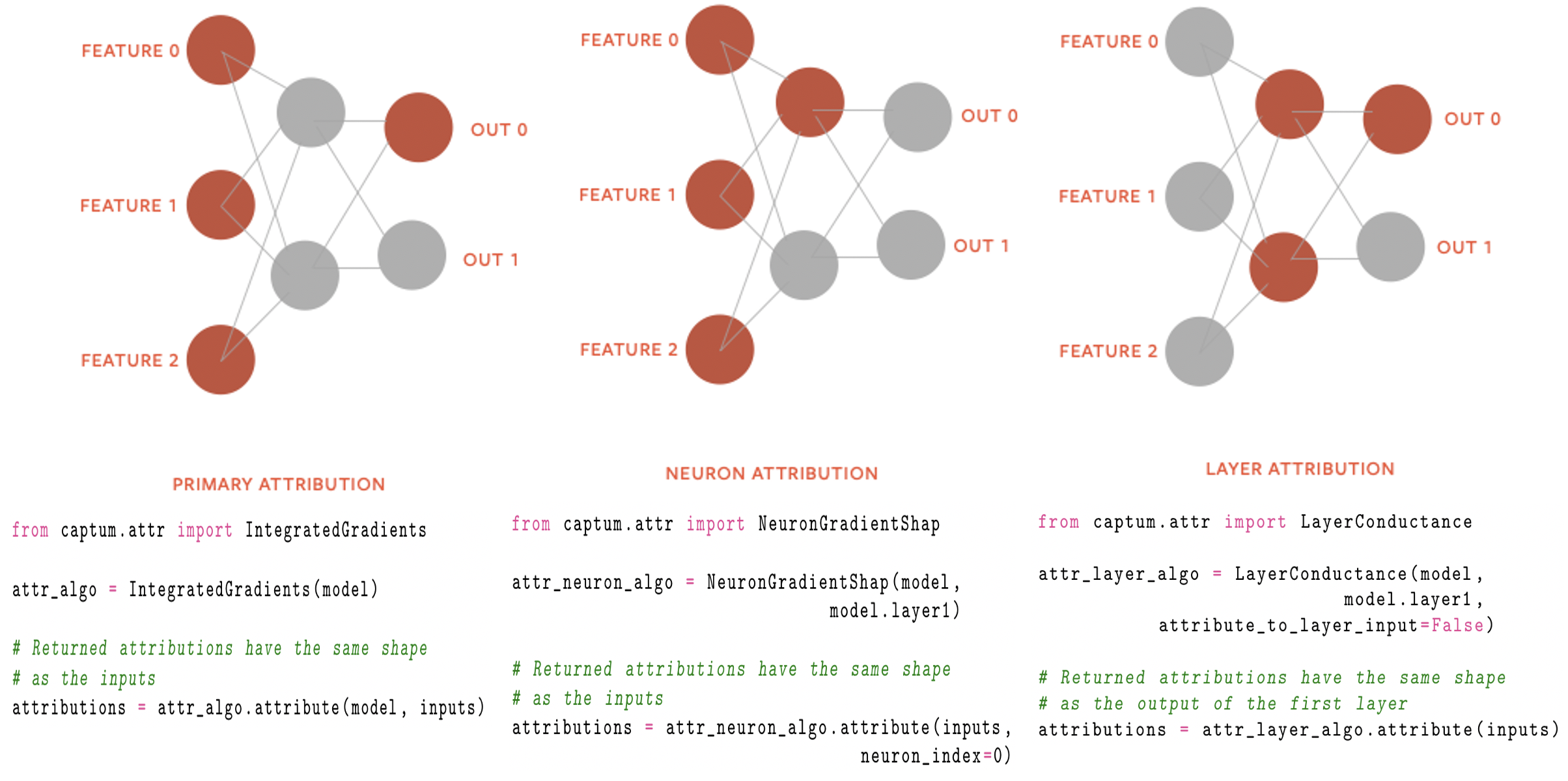}}
\caption{An overview of all three types of attribution variants with example code snippets. The first variant depicted on the far-left side of the diagram represents primary attribution. The algorithms that belong to this group attribute the outputs of the model to its inputs. The middle one allows to attribute internal neurons to the inputs of the model and the one on the right-most side allows to attribute the outputs of the models to all neurons in a hidden layer.}
\label{primary_neuron_layer}
\end{center}
\vskip -0.2in
\end{figure}

The attribution algorithms in Captum can be grouped into three main categories: primary-, neuron- and layer- attributions, as shown in Figure~\ref{primary_neuron_layer}. Primary attribution algorithms are the traditional feature importance algorithms that allow us to attribute output predictions to model inputs. Layer attribution variants allow us to attribute output predictions to all neurons in a hidden layer and neuron attribution methods allow us to attribute an internal, hidden neuron to the inputs of the model. In most cases, both neuron and layer variants are slight modifications of the primary attribution algorithms.

\begin{figure}[ht]
\vskip 0.2in
\begin{center}
\centerline{\includegraphics[width=1.0\columnwidth]{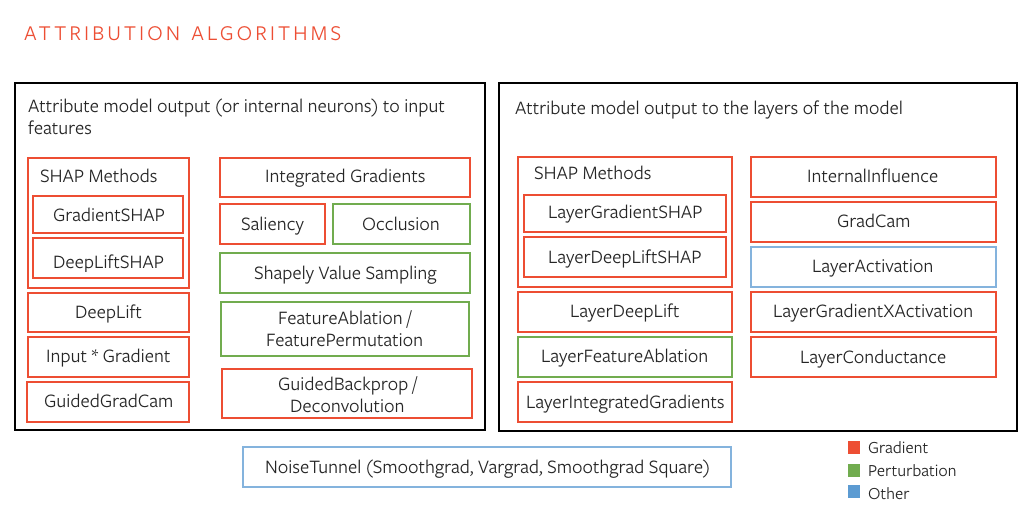}}
\caption{An overview of all the attribution algorithms in Captum. The algorithms grouped on the left side of the diagram are the primary and neuron attribution algorithms. The ones on the right side of the diagram are layer attribution variants. Besides that we can also recognize color-coding of orange for gradient, green for perturbation and blue for algorithms that are neither perturbation nor gradient-based.}
\label{all_attribution_algorithms_png}
\end{center}
\vskip -0.2in
\end{figure}

Most attribution algorithms in Captum can be categorized into gradient and perturbation-based approaches as also depicted in Figure~\ref{all_attribution_algorithms_png}. Some of these algorithms such as GradCam~\cite{gradcam}, GuidedGradCam~\cite{gradcam},
GuidedBackProp~\cite{Springenberg2015StrivingFS}, Deconvolution~\cite{occlusion} and Occlusion~\cite{occlusion} are more popular in the CV community, where they stem from. However, our implementations of those algorithms are generic and can be applied to any model that meets certain requirements dictated by those approaches. For example, GradCam and GuidedGradCam only makes sense for convolutional models.

NoiseTunnel includes generic implementations of SmoothGrad, SmoothGrad Square and VarGrad smoothing techniques proposed in~\cite{smoothgrad}. These methods help to mitigate noise in the attributions and can be used in combination with most attribution algorithms depicted in Figure~\ref{all_attribution_algorithms_png}.

The attribution quality of a number of algorithms such as Integrated Gradients~\cite{ig}, DeepLift~\cite{deeplift}, SHAP variants~\cite{shap}, Feature Ablation and Occlusion~\cite{occlusion} depend on the choice of baseline, also known as reference, that needs to be carefully chosen by the user. Baselines express the absence of some input feature and are an integral part of many feature importance equations. For example, black and white images or the average of those two are common baselines for image classification tasks.

From the implementation and usage perspective, all algorithms follow a unified API and signature. This makes it easy to compare the algorithms and switch from one attribution approach to another. The code snippets in Figure \ref{primary_neuron_layer} demonstrate examples of how to use primary, neuron and layer attribution algorithms in Captum.

\subsection{Scalability}
Many state-of-the-art Neural Networks with a large number of model parameters use large-sized inputs which, ultimately, leads to computationally expensive forward and backward passes. We want to make sure we leverage available memory and CPU/GPU resources efficiently when performing attribution. In order to avoid out of memory situations for certain algorithms, especially the ones with internal input expansion, we slice inputs into smaller chunks, perform the computations sequentially on each chunk and aggregate resulting attributions. This is especially useful for  algorithms such as Integrated Gradients~\cite{ig} and Layer Conductance~\cite{lc} because they internally expand the inputs based on the number of integral approximation steps.
Being able to chunk the inputs into smaller sizes can theoretically allow us to perform integral approximation for infinite number of steps.

In case of feature perturbation, if the inputs are small and we have enough memory resources available, we can perturb multiple features together in one input batch. This requires that we expand the inputs by the number of features that we perturb together and helps to improve runtime performance of all our feature perturbation algorithms.

In addition to this, all algorithms support PyTorch DataParallel, which performs model forward and backward passes simultaneously on multiple GPUs and improves attribution runtime significantly. We made this available for all layer and neuron attribution algorithms, including Layer Activation and NoiseTunnel.

In one of our experiments, we used Integrated Gradients on a pre-trained VGG19 model, for a single 3 x 224 x 224 input image in a GPU environment that has 8 GPUs, each 16GB memory available and gradually increased the number of GPUs while keeping the number of integral approximation steps constant (in this case 2990). From the second column of Table~\ref{ig_distributed} we can see how the execution time decreases substantially as we increase the number of GPUs.
In the second experiment we used the same execution environment, pre-trained model, input image and performed feature ablation by ablating multiple features in one batch using model’s single forward pass. Based on the experimental results shown in third column of Table~\ref{ig_distributed} we can tell that by increasing the number of GPUs from 1 to 8 the execution time drops by approximately 85\%.

\begin{table}[t]
\caption{Runtime performance of Integrated Gradients and Feature ablation for VGG19 with a single input image of shape: 3 x 224 x 224}
\label{ig_distributed}
\vskip 0.15in
\begin{center}
\begin{small}
\begin{sc}
\begin{tabular}{ccc}
\toprule
Number of GPUs & \multicolumn{2}{c}{Execution Time in seconds}  \\ \cline{2-3}
&  Int. Grads & Feature Abl. \\
\midrule
1    & 41.62 & 140.4 \\
4    & 26.41 & 38.39 \\
8    & 15.06 & 22.02 \\
\bottomrule
\end{tabular}
\end{sc}
\end{small}
\end{center}
\vskip -0.1in
\end{table}

\section{Evaluation}
\label{evaluation}
As mentioned in the sections above, both qualitative and quantitative evaluation of feature, neuron and layer importances are challenging research areas.
Human annotations based on visual perception are often subjective, and many quantitative evaluation metrics rely on priors of a dataset domain. We implemented two metrics, infidelity and maximum sensitivity proposed in~\cite{Yeh2019OnT}, in a generic manner for use with any PyTorch model and most Captum algorithms. Similar to other algorithms, we can perform multiple perturbations simultaneously which ultimately helps to improve runtime performance.
Infidelity is the generalization of sensitivity-n~\cite{Ancona2018TowardsBU} metric that relies on the completeness axiom, an important property of many attribution algorithms, that states that the sum of the attributions is equal to the differences of the NN function at its input and baseline.
Given $x \in \mathbbm{R}^N$ as input,  $y \in \mathbbm{R}$ as output, a NN model represented by a continuous function $F: \mathbbm{R}^N \rightarrow \mathbbm{R}$, an attribution function $\Phi: \mathcal{F} \times \mathbbm{R}^N \rightarrow \mathbbm{R}^N$ and a meaningful perturbation $I \in \mathbbm{R}^N$ with a probability measure $\mu_I$, the infidelity of the attribution function $\Phi$ for input sample $x$ and NN predictor function $F$ is defined as:

\begin{equation}
\label{infid}
    INFD_{\mu_{I}}(\Phi, F, x) = \mathbbm{E}_{I \sim \mu_{I}}[(I^T\Phi(F, x) - (F(x) - F(x - I)))^2]
\end{equation}
The formulation~\ref{infid} is for real-valued perturbations $I$ and local attribution functions $\Phi$ that describes the changes of the output when the inputs are slightly perturbed based on the definitions in~\cite{Ancona2018TowardsBU}. Examples of these type of attribution functions are Saliency, GuidedBackprop, Deconvolution, GradCam, GuidedGradCam and DeepLift and Integrated Gradients without multiplying by $x - x_0$, where $x_0 \in \mathbbm{R}^N$ is the baseline chosen for input $x$.

Global attribution functions describe marginal effects of the inputs on the outputs of the model with respect to a chosen baseline~\cite{Ancona2018TowardsBU}. Examples of such attribution functions are Integrated Gradients, DeepLift, SHAP variant of DeepLift and Gradient SHAP aka Expected Gradients~\cite{erion2019learning}. The mathematical formulations of those methods multiply the resulting saliency maps by $x - x_0$.

In case of global attributions, $x_0$ baseline represents the perturbed input corresponding to $x$ and if $x- x_0 \neq 0$, then we can compute the infidelity score without multiplying $\Phi(F, x)$ with the $I^T$ real-valued perturbations. This is because $\Phi(F, x)$ already multiplies the sensitivity scores by $x - x_0$. In this case $I^T$ is reassigned a new role indicating the presence of $x-x_0$ perturbations using a binary representation. Hence, in case of global attributions, $I^T$ represents an indicator function $\mathbbm{I}(i \in S_k)$, where  $S_k \subseteq N$ and $i$ is the $i_{th}$ feature of the $x - x_0$ perturbation.

The other metric, called maximum sensitivity, measures the extent of the attribution change when the input is slightly perturbed. Maximum sensitivity is measured using Monte-Carlo approximation by sampling multiple samples from an $L_p$ ball using a perturbation radius $r \in \mathbbm{R}$. It is a more robust variant of Lipschitz Continuity metric~\cite{AlvarezMelis2018OnTR} and does not require the attribution function to be continuous everywhere.
The users of Captum can use any perturbation function that computes perturbations within any $L_p$ ball. We provide a default implementation that samples uniformly from $L_{\infty}$ ball.

Given $x \in \mathbbm{R}^N$ as input, $y \in \mathbbm{R}^N$ as perturbed input, a NN model represented by a continuous function $F:\mathbbm{R}^N \rightarrow \mathbbm{R}$, an attribution function $\Phi: \mathcal{F} \times \mathbbm{R}^N \rightarrow \mathbbm{R}^N$ and perturbation radius $r \in \mathbbm{R}$, maximum sensitivity is defined as:
\begin{equation}
\label{max_sens}    
    SENS_{MAX}(\Phi, F, x, r) = \mathop{max}_{||y - x|| \leq r}\frac{||\Phi(F, y) - \Phi(F, x)||}{||\Phi(F,x)||}
\end{equation}
In the next section we demonstrate results of these two metrics in different applications.

\section{Applications}
\label{applications}
\begin{figure}[ht]
\vskip 0.2in
\begin{center}
\centerline{\includegraphics[width=1.0\columnwidth]{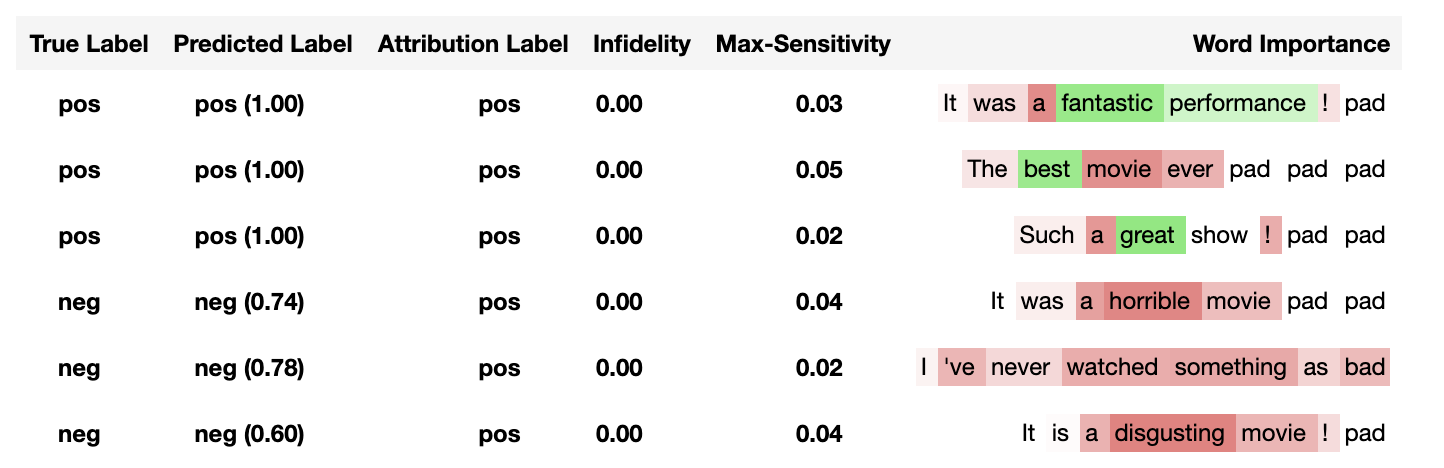}}
\caption{Visualizing salient tokens computed by integrated gradients that contribute to the predicted class using a binary classification model trained on IMDB dataset. Green means that those tokens pull towards and red that they pull away from the predicted class. The intensity of the color signifies the magnitude of the signal.}
\label{text_classif_infid_sens}
\end{center}
\vskip -0.2in
\end{figure}

The generality of Captum library allows us to apply it to different types of NNs. One of the commonly used applications is text classification. We used a pre-trained text classification model on IMDB dataset~\cite{maas-EtAl:2011:ACL-HLT2011} and Integrated Gradients to identify most salient tokens in the text. Figure \ref{text_classif_infid_sens} visualizes color-coded contributions of each token to the output predictions. We also computed infidelity and max-sensitivity scores. The perturbations for infidelity metric are sampled uniformly from a normal distribution $\mathbbm{N}(0, 0.03)$ and for max-sensitivity from a $L_\infty$ ball with a radius of 0.03. The norm $||.||$ used in max-sensitivity computations is the $L_2$ norm.

\begin{figure}[ht]
\vskip 0.2in
\begin{center}
\centerline{\includegraphics[width=1.0\columnwidth]{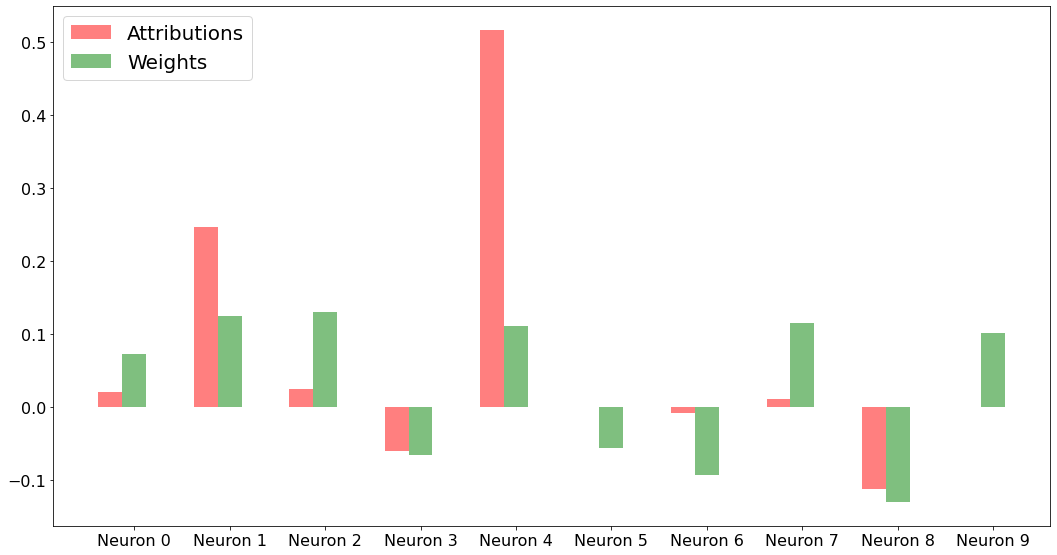}}
\caption{Visualizing normalized attribution scores and weights for all ten neurons in the last linear layer of a simple four MLP model trained on Boston house prices dataset.}
\label{regression}
\end{center}
\vskip -0.2in
\end{figure}

\begin{figure}[ht]
\vskip 0.2in
\begin{center}
\centerline{\includegraphics[width=1.0\columnwidth]{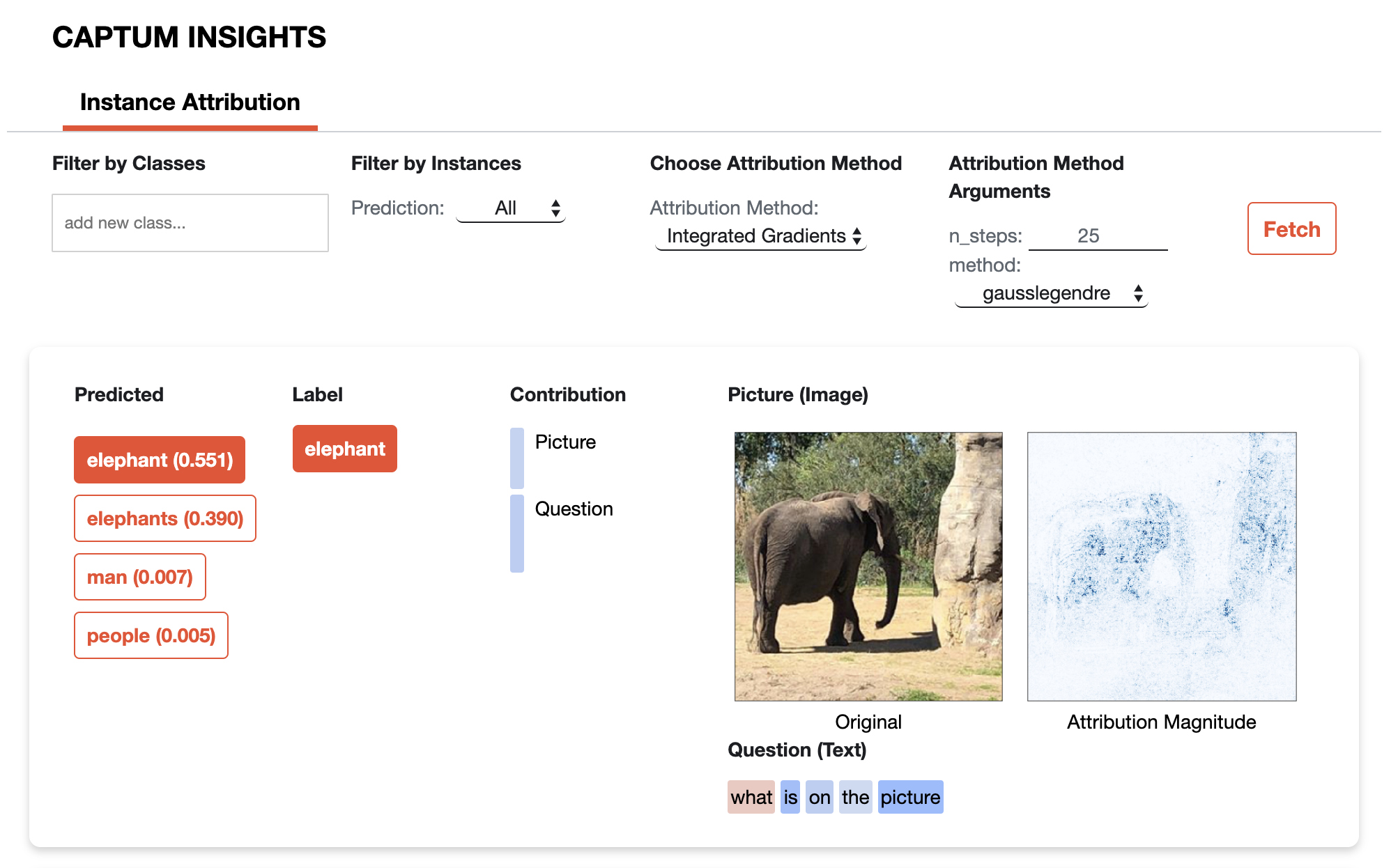}}
\caption{Captum Insights interactive visualization tool after applying integrated gradients to the Visual Question Answering multi-modal model. The tool also visualizes aggregated attribution magnitudes of each modality.}
\label{multimo}
\end{center}
\vskip -0.2in
\end{figure}

Classification models are not the only types of models that Captum supports. As an example, we built a regression model using Boston house prices dataset~\cite{HARRISON197881} and a simple four layer NN using linear layers and ReLUs. We attributed the output predictions to the last linear layer using layer conductance algorithm and plotted them together with the learned weights of that same layer as shown in Figure~\ref{regression}. Here we can see that the weights and the attribution scores are aligned with each other. Both scores were normalized using $L_1$ norm.

In order to improve model debugging experience, we developed an interactive visualization tool called Captum Insights. The tool allows to sub-sample input examples and interactively attribute different output classes to the inputs of the model using different types of attribution algorithms depicted in Figure~\ref{multimo}.

\subsection{Multimodality}
\label{multimodal}
The support for multi-modal Neural Networks is one of the core motivations of Captum library. This allows us to apply Captum to machine-learning models that are built using features stemming from different sources such as audio, video, image, text, categorical or dense features. Aggregated feature importance scores for each input modality can reveal which modalities are most impactful. In the example of attributions computed for a multi-modal Visual Question Answering model depicted in Figure~\ref{multimo}, we can tell whether the stronger predictive signal is coming from text or image.

\section{Conclusion}
\label{conclusion}
We presented a unified model interpretability library for PyTorch, called Captum, that supports generic implementations of a number of gradient and perturbation -based attribution algorithms. Captum can be applied to NN models of any type and used both in research and production environments. Furthermore, we described how the computations are scaled and how we handle large-sized inputs. In addition to that, we also added support for two generic quantitative evaluation metrics called infidelity and maximum-sensitivity. 
Lastly, we walked through different types of applications including multi-modal models and introduced Captum Insights model debugging tool.

\section{Future Work}
\label{future_work}
Our future work involves both expanding the list of attribution algorithms and looking beyond attribution methods for model understanding. Beyond feature, neuron and layer attribution we are also looking into adversarial robustness and the intersection between these two fields of research. Concept-based model interpretability that aims to explain the models globally using human understandable concepts is another interesting direction to explore. Besides that, visualizing high-dimensional embedding vectors in the latent layers and being able to debug the models and understand what information a single or a group of neurons in the layers encode are other interesting avenues to explore.

\section{Acknowledgements}
\label{acknowledgements}
We are grateful to all Captum contributors and users both external and internal, and PyTorch community for their support, contributions and feedback. Their input and contributions helped us to create a self-contained model understanding library that can be used both in research and production. We would like to say thank you to Soumith Chintala, Joe Spisak, Alban Desmaison and Francisco Massa from core PyTorch team for supporting us with Open Source and PyTorch related questions. Davide Testuggine for his support and initial discussions on model interpretability and integrated gradients. We would also like to thank Tucker Hart for implementing the initial prototype of Captum Insights, Jessica Lin for helping us with the documentation and Fuchun Peng for reviewing this paper and providing valuable feedback.

\bibliographystyle{plain}
\bibliography{main}
\end{document}